# Reduce Computational Complexity for Convolutional Layers by Skipping Zeros

Zhiyi Zhang, Pengfei Zhang, Zhuopin Xu, Qi Wang

*Hefei Institutes of Physical Science, Chinese Academy of Sciences,* Hefei, Anhui, China
*University of Science and Technology of China,* Hefei, Anhui, China
gilgamesh@mail.ustc.edu.cn, pfzhang@aiofm.ac.cn, xuzp@iim.ac.cn, wangqi@ipp.ac.cn

*Abstract*—Convolutional neural networks necessitate good algorithms to reduce complexity, and sufficient utilization of parallel processors for acceleration. Within convolutional layers, there are three types of operators: convolution used in forward propagation, deconvolution and dilated-convolution utilized in backward propagation. During the execution of these operators, zeros are typically added to tensors, leading to redundant calculations and unnecessary strain on hardware. To circumvent these inefficiencies, we propose the C-K-S algorithm, accompanied by efficient GPU implementations. C-K-S trims filters to exclude zero-padding. For deconvolution and dilated-convolution, C-K-S transforms sparse tensors into dense tensors, and standardizes the local computational rules to simplify the hardware control. The experimental results demonstrate that C-K-S offers good performance in terms of speed and convergence, surpassing the capabilities of PyTorch and cuDNN in certain scenarios.

*Keywords*—convolutional neural networks, deconvolution, dilated convolution, graphic processing units (GPU).

## I. INTRODUCTION

Convolutional Neural Networks (CNNs) have found extensive application across various domains, yielding remarkable achievements. CNNs [2]-[6] are neural networks with convolutional layers (conv-layer). The propagation of conv-layers involves three distinct convolutional operators: convolution, deconvolution (transposed-convolution), and dilated-convolution. Convolution generates outputs in the forward propagation, whereas the other two operators calculate the gradients in backward propagation. These three operators are main speed bottlenecks for training CNNs, due to their high time-complexity and resource occupancy.

Effective implementations [7]-[15] of convolutional operators are generally based on parallel processors such as GPU and FPGA, since they offer efficient execution of the parallelizable dot-products derived from these operators. In the realm of convolutional operators design, it's crucial to consider algorithms, which has low complexity and can be easily adapted to hardware implementations.

When performing convolutional operators, it's common to include zero-elements (0s) in tensors. This is typically done to pad feature-maps, or create sparse tensors for deconvolution and dilated-convolution. However, these 0s lead to redundant calculations (0-calculation), as 0 has no effect on multiplication. On the other hand, hardware needs supplementary resources to handle these 0s, resulting in a decrease in overall performance. By skipping these 0s, it's possible to reduce the computational complexity, and improve hardware efficiency.

To mitigate the inefficiencies caused by 0s, we present the C-K-S algorithms, with explicit formulas and pseudo code. The C-K-S algorithm consists of three components: ConvV2, KS-deconv, and Sk-dilated. The filter-trimming of ConvV2 provides a constant factor acceleration, by excluding all padded 0s. Through filter-reconstruction and leaping-access, KS-deconv and Sk-dilated convert sparse tensors to dense tensors, enabling $stride^N$ and $dilate^N$ times acceleration for N-dimensional deconvolution and dilated-convolution.

In this work, we provide efficient GPU implementations of C-K-S, and further integrated them into Dragon-Alpha [15][21], a tensor computing framework that we developed. C-K-S is founded on math transformations instead of specific systems and architectures, so it's not limited to GPU.

We analyze the performance of C-K-S, by benchmarking it against cuDNN [9] and PyTorch [13]. The experimental results indicate that, C-K-S is effective in accelerating convolutional operators, especially for small feature-maps. Besides, C-K-S shows stable convergence, when training CNNs on Cifar10 [24] and ILSVRC2012 [25].

## II. BACKGROUND

### A. Convolutional Operators

Convolutional operators are operators that perform convolution and its variations. Their input and output matrices are respectively referred as input-feature-maps and output-feature-maps. The filters (convolutional kernel) are tensors, that act on input-feature-maps to generate the output-feature-maps.

Convolution, deconvolution, and dilated-convolution are the three most commonly used convolutional operators, which are also essential components of convolutional and deconvolutional layers. Conv-layers employ convolution in forward propagation, and use deconvolution in backward propagation. Conversely, deconvolutional layers employ deconvolution in forward propagation, and convolution in backward propagation. In both types of layers, dilated-convolution is used to find the gradients of filters.

Convolution and deconvolution are a pair of inverse operators. Convolution is commonly used in down-sampling, whereas deconvolution is frequently utilized to expand feature-maps. Convolution extracts more profound semantic information from shallower levels, while deconvolution decodes higher-level features to lower-levels. Given their characteristics, convolution is well-suited for classification

This research was supported by National Natural Science Foundation of China (No.32070399); Anhui Provincial Key Research and Development Program (No.2023n06020016, No.2023n06020028), Anhui Science and Technology Major Project (No.202103a06020014), the Chinese Academy of Sciences-Henan Province Achievement Transfer and Transformation Project (No. 2022208).

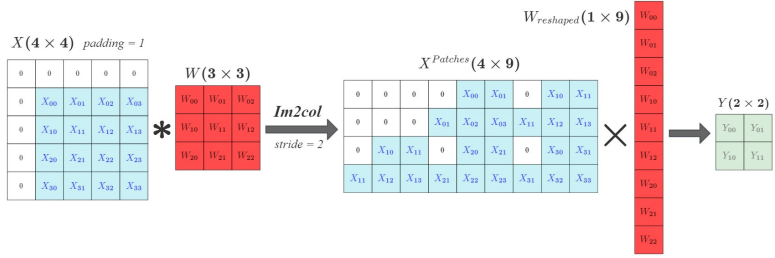

Fig. 1. Through *im2col*, the 2D-convolution $X * W$ cam be transformed to a matrix-multiply $X^{Patches} \times W^{reshaped}$.

and recognition, while deconvolution is typically used in generation and segmentation.

*B. Approaches to Implement 2D-Convolution on GPUs*

The approach of Krizhevsky [8] is to compute the convolution directly. It's very efficiency in some cases, but may not perform well in others, and requires specialized implementations for the corner cases of convolution.

The Fast-Fourier-Transform (FFT) convolution exhibits a significantly smaller time-complexity compared to dense convolution. However, this approach has several drawbacks. FFT is less efficiently in case of small filters or large stride, and requires temporary memory to pad filters the same size as input-feature-maps. Besides, FFT can't fully utilize GPUs, due to its recursive nature and low computational density. Hence, its performance is more remarkably influenced by memory-access rather than computing-speed, despite the latter factor being the biggest advantage of GPUs.

Caffe [12] lowers 2D-convolution to matrix-multiply by *im2col*, as shown in Figure 1. This approach is fast and robust, but it requires auxiliary memory to store the unfolded tensors. CuDNN [9] has optimized this method, by implicitly integrating *im2col* to GEMM (General-Matrix-Multiply), resulting in higher speed without auxiliary memory. The GEMM-based approaches are well-suited for GPUs. Matrix-multiply consists of parallelizable dot-products, allowing GPUs to effectively hide the latency of memory-access. The overlapping of continuous memory-access occurs in these dot-products, enabling lower cost of memory-access through the use of shared-memory and vector-datatypes. Besides, GPUs can fuse multiplications and additions to multiply-accumulate operations. This operation sacrifices a bit of accuracy, but it sensibly quicker than regular multiplication.

The direct [8] and implicit-GEMM [9] approaches are referred to implement C-K-S on GPUs. Since deconvolution and dilated-convolution are variations of convolution, it's possible to implement them using a similar methodology.

*C. Related Works of Deconvolution*

The main work of deconvolution acceleration is to avoid 0-caluclations, and optimize hardware control.

Orosa *et. al.* [17] devise a compile-time computation scheduling technique, aiming to eliminate 0-calculations with minimal overhead. Cutlass [10] offers implementations of *strided_dgrad*, which skips 0-calculations through thread indices of CUDA. Chang *et. al.* [18] present the TDC method accompanied by FPGA implementations, which converts sparse deconvolution to dense-convolution. In the study of Chang *et. al.* [19], a kernel-decomposition method to exclude 0-caluclations, has been implemented based on TSMC-40nm-CMOS-technology. Vadakkeveedu *et.al.* [20] also implement a kernel-decomposition method for mixed-precision (float16 and float32) based on TensorFlow [11].

Our KS-deconv may have a border range of applicability compared to the methods in studies [10][17], as it doesn't relies on the architecture of CUDA or specific compilers.

In comparison to the TDC [18] method, KS-deconv has a lower time-complexity, owing to its capability to decompose filters to smaller kernels.

The decomposition outcomes in researches [19]-[20] are similar to KS-deconv's in certain scenarios. However, these studies don't provide explicit math-formulas or algorithms, and are based on systems different than ours. These factors impede us to replicate their algorithms and make general comparisons. Besides, these works mainly focus on the analysis in terms of speed, potentially neglecting discussions about convergence.

To achieve higher performance, we further combined filter-trimming with KS-deconv in this work.

## III. MOTIVATIONS

The main goal of C-K-S is to minimize the additional intricacy that arises from including 0s in convolutional operators, particularly in terms of time and hardware control. To ensure its broad applicability, the fundamental principle of C-K-S should be rooted in mathematical tensor operations, rather than specific systems or architectures. These mathematical operations should be simple and have minimal interdependence, in order to align with the nature of SIMD (single instruction, multiple data).

For clear description, the notations of 2-dimensional (2D) conv-layers are listed in Table I, and some relevant math notations are presented in Table II.

TABLE I. PARAMETER NOTATIONS OF CONV2D-LAYERS

| Notation | Explanation |
|---|---|
| $X_{n, ih, iw, ic}$ | An indexed element of input-feature-maps |
| $Y_{n, oh, ow, ic}$ | An indexed element of output-feature maps |
| $W_{oc, fh, fw, ic}$ | An indexed element of filters |
| $I_H \setminus I_W$ | The height \ width of $X$ |
| $O_H \setminus O_W$ | The height \ width of $Y$ |
| $F_H \setminus F_W$ | The height \ width of $W$ |
| $N \setminus I_C \setminus O_C$ | The batch-size \ input-channels \ output-channels |
| $sh \setminus sw$ | The stride on height \ width axis |
| $ph \setminus pw$ | The padding on height \ width axis |

TABLE II. MATH NOTATIONS

| Notation | Explanation |
|---|---|
| $*$ | Convolution or Scalar-multiply |
| $\times$ | Cartesian-product or matrix-multiply |
| $\odot$ | Element-wise multiply |
| $\langle \cdot \rangle$ | Matrix or Vector |
| in-range-of | Return true if an index is within the range of a tensor, otherwise return false |
| $O(f(n))$ | Represents the asymptotic time complexity, where $n$ refers the size of data |

Convolutional operators are composed of multiplication and accumulation operations, that are applied to both the filters and input-feature-maps. Among these operations, only the operations acting on non-zero elements are necessary, while the rest are otiose.

To produce output-feature-maps with specified size using convolutional operators, it's usually necessary to pad certain 0s on the boundary of input-feature-maps. In many CNNs, the feature-maps become smaller with the increase of depth due to down-sampling, while the padding size remains unchanged. This leads to small proportions of padded 0s in input-feature-maps at shallow depths, but potentially large proportions in deep layers, resulting in a non-negligible number of 0-calculations. Figure 2 illustrates this phenomenon in the case of 2D-convolution, where both proportions become bigger, as the size of the square input-feature-maps decreases.

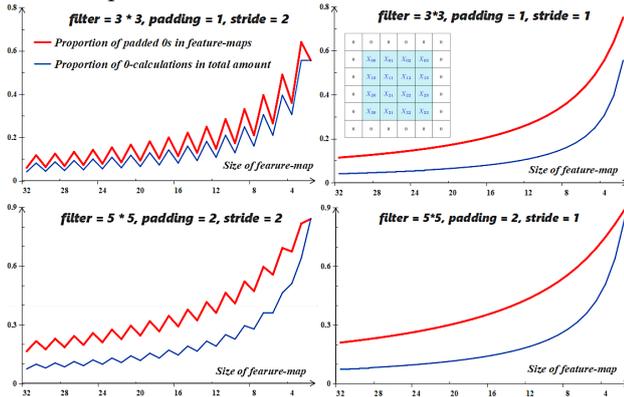

**Fig. 2.** Padding causes more 0-calulations on smaller feature-maps.

In backward propagation of conv-layers with a non-unit stride (greater than 1), $(stride - 1)$ 0s are inserted between adjacent elements of $\nabla Y$, leading to the sparsity. After that, $\nabla Y$ serves as filters in dilated-convolution to find $\nabla W$, and as feature-maps in deconvolution to find $\nabla X$. Typically, the inserted 0s make up at least 50% of $\nabla Y$, meaning that the majority of the calculations are 0-calculations. As depicted in Figure 3, in the context of deconvolution and dilated-convolution, the proportions of non-0-calculations rapidly decrease with the increase of stride, finally approaching 0.

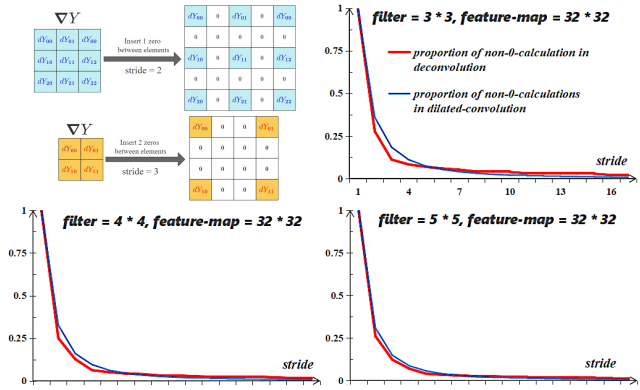

**Fig. 3.** Zero-insertion causes more 0-calculations with bigger stride.

Among the three convolutional operators of conv-layers, convolution is the most straightforward to accelerate, because the tensors involved are dense, and the local calculation rules remain consistent. The second easiest is dilated-convolution, where the filters are sparse but the input-feature-maps are dense, allowing for the consistency of local calculation rules. The most challenging one is deconvolution. In the presence of non-unit stride, its input-feature-maps become sparse, and the distribution of inserted 0s varies across different patches of the input-feature-maps, leading to heterogeneous local computing rules.

Typically, the memory of parallel processors is high bandwidth but low capacity, making it a limited resources. As a result, their processes of zero padding or insertion are usually achieved through conditional-statements. Namely, when fetching a particular element, return its value if its index meets certain conditions, but 0 otherwise. Such logical implementation doesn't require temporary memory, but sacrifices certain speed as it disrupts SIMD. To alleviate the negative effect, it's imperative to streamline the control of processing 0s, and minimize the reliance of conditional statements.

To exclude the padded 0s, ConvV2 applies a trimming technique to filters, which also removes the conditional statements caused by zero-padding.

For deconvolution and dilated-convolution, KS-deconv and Sk-dilated transforms sparse tensor $\nabla Y$ to dense tensor, to eliminate the inserted 0s. Additionally, KS-deconv homogenizes the local computational rules of sparse deconvolution, resulting in simplified control and enhanced ease of acceleration.

## IV. ALGORITHMS AND IMPLEMENTATIONS

This section explains the 2D algorithms and implementations of C-K-S. The pseudo code of algorithms is listed in Appendix, which can help with comprehension.

As the conv2D-layers of DNNs involve channels and batches, their filters $W \subset \mathbb{R}^{O_C \times F_H \times F_W \times I_C}$, input-feature-maps $X \subset \mathbb{R}^{N \times I_H \times I_W \times I_C}$, and output-feature-maps $Y \subset \mathbb{R}^{N \times O_H \times O_W \times O_C}$ are all 4D tensors. The forward propagation of conv2D-layers is denoted by (1), while the backward propagation is represented as (2) and (3).

$$Y = conv_{2D}(X, W) \quad (1)$$
$$\nabla X = deconv_{2D}(\nabla Y, W^{rot180}) \quad (2)$$
$$\nabla W = dilated\_conv_{2D}(X, \nabla Y) \quad (3)$$

ConvV2 is discussed in terms of forward propagation, while KS-deconv and Sk-dilated are explained in a view of backward prorogation.

### A. ConvV2: Convolution with Trimmed Filters

When performing 2D convolution, the filters $W$ move across the input feature-map $X$, which is zero padded. The filters then perform dot-products with each covered portion of $X$, referred to as patches. Within each patch, the padded 0s only present at the boundary, while the non-zero elements are located in the center, forming a rectangular shape. Clearly, only the non-zero elements within these rectangles, and their corresponding region in $W$, can yield meaningful calculations of convolution.

To ensure that only the non-zero rectangle of each patch is considered, Conv2 trims $W$ to match this rectangle shape, starting from the top-left corner $(fh_s, fw_s)$, and ending at the bottom-right corner $(fh_e - 1, fw_e - 1)$. The height and width of these two same-size non-zero rectangles are $(fh_e - fh_s)$ and $(fw_e - fw_s)$ respectively. Figure 4 presents an example of ConvV2, where the filter is adjusted for the four patches. The time-complexity of ConvV2 is 50, smaller than the 72 of normal convolution shown in Figure 1.

In our implementations, the filter-trimming is achieved by moving pointers and restricting memory-access, which is low-cost and doesn't necessitate auxiliary memory. The integration of filter-trimming into convolutional operators, only requires few lines of code, typically less than twenty. In addition to reducing complexity, filter-trimming eliminates the need for conditional statements when loading elements, so that certain index-solving calculations can be omitted. Our implementations ensure that all threads in the same block have the same trimmed-filters, so there exists no warp-divergence that disrupts SIMD. As GPUs don't need additional registers to handle warp-divergence, ConvV2 kernel-functions uses less than 128 registers per thread, usually fewer than those of implicit-GEMM, enabling more active-threads on GPUs.

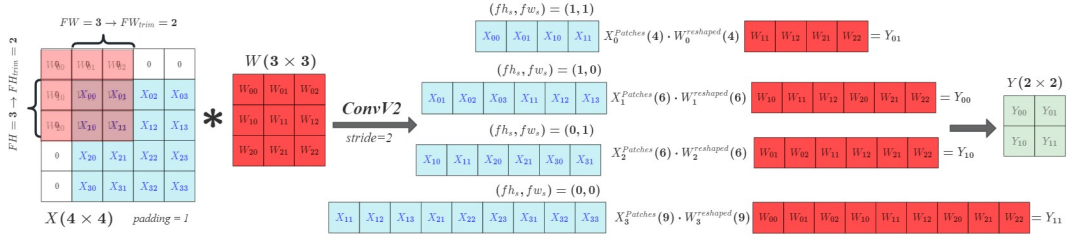

Fig. 4. ConvV2 trims filters to exclude all padded 0s.

When the size of $X$ is sufficiently large, to improve memory bandwidth, the dimension order of $W$ will be transposed from $(O_C \times F_H \times F_W \times I_C)$ to $(F_H \times F_W \times I_C \times O_C)$ with $\Theta(n)$ complexity. This filter-transposition incurs about 6% increase in overall speed. The temporary memory used to store the transposed $W$ will be released once the convolution is completed.

*B. KS-deconv: Kernel-Split Deconvolution*

KS-deconv is a deconvolution algorithm, specifically designed to handle the sparsity in non-unit stride scenarios.

For sparse 2D deconvolution, the input-feature-maps contain $(stride - 1)$ 0s between adjacent non-zero elements. The distribution of inserted 0s varies among different patches; however, it's possible to categorize these 0-distrbutions into $(sh * sw)$ classes across all patches. Each individual element in the output-feature-maps corresponds to a specific patch of the input-feature-maps. The remainder of this element's coordinate, relative to stride, can also be classified into $(sh * sw)$ classes. It has been observed that, there is a surjection between the classes of these remainders and 0-distributions, meaning that these 0-distributions can be distinguished by their corresponding remainders. Besides, each element of the output-feature-maps can be calculated, through a dot-product on two collections of non-zero elements. One collection is derived from the non-zero segmentation of a specific patch, while the other is obtained from the segmentation of filters corresponding to this patch. Such two collections can be generated by selecting elements from appropriate positions with $\langle sh\ sw \rangle$ step-size.

Consequently, 2D deconvolution can be achieved by splitting the filters into $(sh * sw)$ parts, and then applying them to corresponding subsets of input-feature-maps. The local computational rules remain consistent within each subset. That's the fundamental idea of KS-deconv.

KS-deconv is composed of three stages. In the backward propagation of conv2D-layers, $\nabla Y$ acts as input-feature-maps of deconvolution, while $\nabla X$ serves as the output-feature-maps. Figure 5 illustrates KS-deconv with an example.

**Stage1** *kernel-split*: The filters $W$ are rotated 180 degrees, and then split to construct $(sh * sw)$ smaller kernels. The shape of the $(y, x)_{th}$ smaller kernel $C_{y,x}$ is $(O_C \times \lceil \frac{F_H - y}{sh} \rceil \times \lceil \frac{F_W - x}{sw} \rceil \times I_C)$. Finally, all smaller kernels are concatenated into a 6D tensor $C$, which has continuous memory space.

**Stage2** *stride-1-convolutions*: All smaller kernels are used to perform unit-stride convolutions with $\nabla Y$, and eventually $(sh * sw)$ outputs are generated. The convolution with $C_{y,x}$ has a zero-padding of $\langle oph_{y,x}\ opw_{y,x} \rangle$ size.

**Stage3** *composition*: All outputs of Stage2 are composed to obtain $\nabla X$. Each output is assigned a start-point $(ih_s, iw_s)$ to write results to $\nabla X$. The start-points are adjusted to non-negative values, and each coordinate $(ih, iw)$ of a specific result is checked to ensure it falls within the range of $\nabla X$. When $\langle I_H\ I_W \rangle$ is an integral multiple of $\langle sh\ sw \rangle$, there is no need to check $(ih, iw)$, resulting in improved efficiency due to less conditional statements.

The time-complexity of KS-deconv is $\Theta(n^3)$, and that of the common approach is $\Theta(sh * sw * n^3)$, which is $(sh * sw)$ times of the first. As the stride increases, the gap between the two grows rapidly.

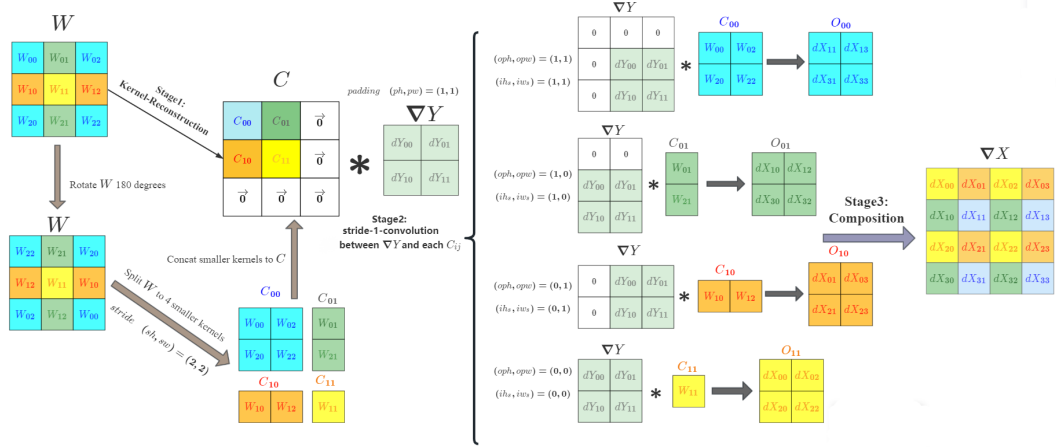

Fig. 5. Three stages of KS-deconv. In Stage1, the filter $W(3 \times 3)$ is rotated first, then split to 4 smaller kernels $C_{00}(2 \times 2)$, $C_{01}(2 \times 1)$, $C_{10}(2 \times 1)$ and $C_{11}(1 \times 1)$, as the stride is 2. These smaller kernels are concatenated to $C(2 \times 2 \times 2 \times 2)$ which has continuous memory space. In Stage2, $C_{00}$, $C_{01}$, $C_{10}$ and $C_{11}$ are respectively used to perform unit-stride convolution with $\nabla Y$, and the outputs are $O_{00}$, $O_{01}$, $O_{10}$ and $O_{11}$. In Stage3, $O_{00}$, $O_{01}$, $O_{10}$ and $O_{11}$ are composed to obtain $\nabla X$.

In our implementations, Stage2 and Stage3 are fused, enabling Stage3 to directly use the outputs of Stage2 in registers. The cost of Stage1 is low, and can be disregarded when feature-maps are sufficiently large. In most cases, the auxiliary memory required by $C$ is significantly lower than that required by feature-maps, and it will be freed once the deconvolution is completed. Besides, the complexity $\Theta(n)$ of Stage1 can be regard as a low-order-infinity of the complexity $\Theta(n^3)$ of Stage2. We have also tried to fuse all three stages into a single operator. This approach doesn't require auxiliary memory, because Stage1 is accomplished through index-conversion. However, it's slower than the aforementioned method, since its memory-access is less continuous, and it has duplicate index-calculations, thereby increasing overhead.

The filter-trimming has been integrated into Stage2 of KS-deconv, developing KS-deconv-V2. As depicted in Figure 6, the time-complexity of Stage2 in KS-deconv-V2 is 50, less than the 72 observed in KS-deconv (Figure 5).

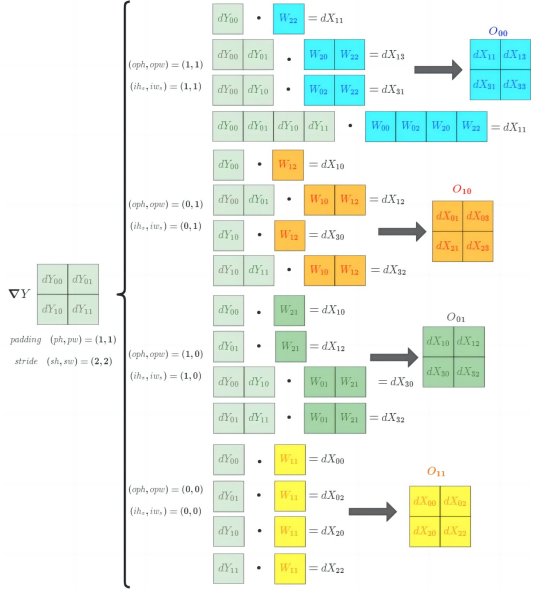

**Fig. 6.** Stage2 of KS-deconvV2, with filter-trimming.

C. *Sk-dilated: Skip 0s in Dilated-Convolution*

In dilated-convolution, the filters are sparse with $(dilate - 1)$ 0s inserted between adjacent elements. The non-zero elements in the filters correspond to indices on the height and width axes, which are integral multiples of dilate. Therefore, the inserted zeros can be skipped, by fetching elements in the filters and input-feature-maps with a leaping step-size equal to dilate.

Sk-dilated has been introduced as an alternative approach to dilated-convolution. In this approach, 0s are not inserted into the filters. Instead, within each dot-product of dilated-convolution, elements in the filters are fetched with unit-step-size, while elements in the input-feature-maps are fetched with a leaping step-size equal to dilate.

In the backward propagation of a particular conv2D-layer, dilated-convolution uses $\nabla Y$ as filters, and $X$ as input-feature-maps. The value of dilate is equal to the stride $\langle sh\ sw \rangle$ of the conv-layer. As shown in Figure 7, within the current patch, Sk-dilated fetches elements based on the following 2D-index sequence: $(0, 0) \to (0, 2) \to (2, 0) \to (2, 2)$.

The time-complexity of Sk-dilated is $\Theta(n^3)$, and that of the common approach is $\Theta(sh * sw * n^3)$, which is $(sh * sw)$ times of the first. With the increase of stride, the gap between the two rapidly grows.

Through *im2col*, Sk-dilated-convolution can be lowered to a matrix-multiply involving $A \subset \mathbb{R}^{G_N \times G_K}$ and $B \subset \mathbb{R}^{G_K \times G_M}$. On GPUs, the calculations of Sk-dilated are distributed among a specific number of thread-blocks, based on the sizes of $A$ and $B$. This number determines the parallelism, and affects the overall speed. When using Sk-dilated in the backward propagation of conv2D-layers, we have $G_N = O_C$, $G_M = F_H * F_W * I_C$, and $G_K = N * O_H * O_W$. If the number of thread-blocks is solely decided by $G_N$ and $G_M$, the parallelism may be insufficient, since these values don't increase with the size of input-feature-maps. To improve the parallelism, the map-reduce is employed: $A$ and $B$ are split to $G_Z$ segments along $G_K$ axis, the calculations are then performed on these segments concurrently, and the results obtained from each segments are aggregated to yield the final results. Because the calculations on these segments contribute the most time-complexity, the parallelism becomes about $G_Z$ times that of the original.

To balance parallelism and memory usage, $G_Z$ should be carefully chosen. In a specific conv-layer, a convolution corresponds to a deconvolution and a dilated-convolution. There operators share the same parameters of the conv-layer, and their number of thread-blocks are respectively $N_\alpha$, $N_\beta$ and $N_\gamma$. As $N_\alpha$ and $N_\beta$ increase with the feature-map-size of the conv-layer, $G_Z$ can be positive related to $(N_\alpha + N_\beta)/N_\gamma$. To restrict $G_Z$, there also needs a lower-bound and a upper-bound. The upper-bound can be decided by the number of streaming multi-processors.

The filter-trimming has been integrated into Sk-dilated, resulting the development of Sk-dilated-V2. As shown in Figure 8, the time-complexity of Sk-dilated-V2 is 50, lower than the 72 observed in Sk-dilated (Figure 7).

D. *Optimizations for GPU*

Our implementations of C-K-S utilize CUDA C++ at low-level, and Java at top-level. More than 100 CUDA kernel-functions have been developed for C-K-S. Some of these functions are general solutions to ensure the lower limit of performance, while some are specially tailored for higher performance in certain scenarios.

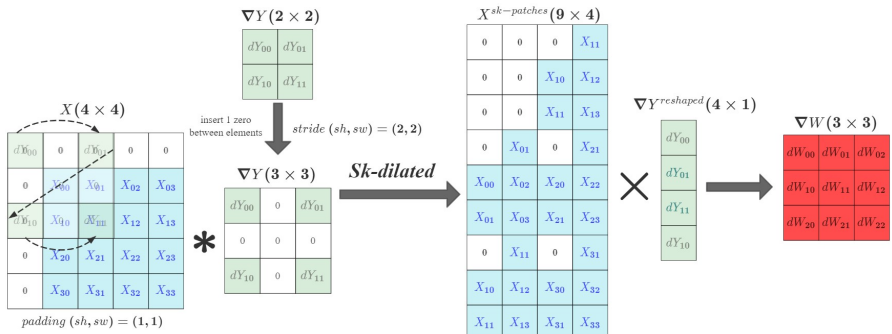

**Fig. 7.** Sk-dilated fetches elements in leaping-steps.

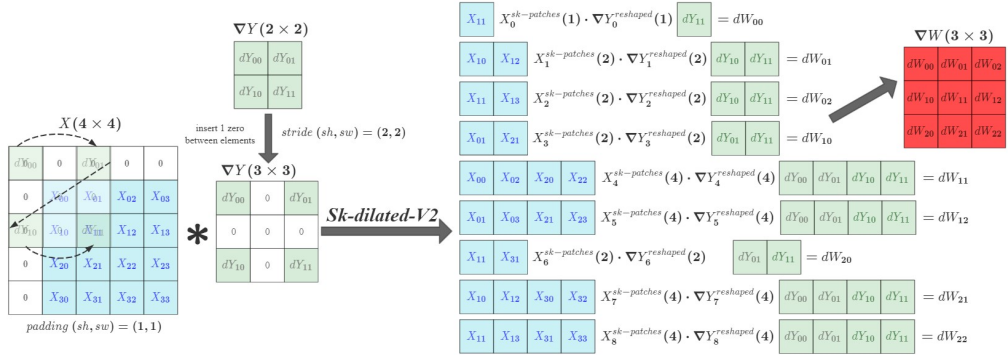

**Fig. 8.** Sk-dilated-V2: padded 0s are excluded. The time-complexity is 50, less than the 72 of *Sk-dilated* (Fig. 8).

The following tricks are used to optimize kernel-functions:

*1) Minimize the latency of memory-access:* The cost of memory-access is reduced by using double-buffered shared-memory, and improving the re-usability of elements. To hide the latency, the heaviest kernel-functions execute a big batch of instructions ($8*8*16$ multiply-add operations) per round. The last dimensions of tensors are implicitly padded to multiples of 4, and some memory-operations are merged using vector-datatypes, so that 128bit can be more used as a unit of memory access to improve the bandwidth.

*2) Minimize the use of integral division and remainder operations:* These operations are much slower than the floating-point operations on GPUs. However, convolutional kernel-functions require these operations to calculate indices, which lowers the performance. We try to replace them with bitwise operations. For specific sizes of the 2D filters, such as $(3\times3)$, $(4\times4)$, $(5\times5)$ and $(7\times7)$, the indices are pre-calculated and stored in constant-memory to maximize broadcasting. The pre-calculated indices are then retrieved by kernel-functions.

*3) Minimize the use of registers:* Most kernel functions use less than 129 registers per thread to ensure parallelism. Typically, 128 serves as a critical threshold that impacts the number of active-threads on GPUs.

*4) Minimize the use of conditional statements:* Some conditional statements are substituted with equivalent tables and expressions. Besides, expanding certain loops can reduce the number of conditional statements.

Further enhancement can be achieved at assembly level, by integrating ptx-code and modifying sass-code of CUDA. However, these works requires huge time and expertise in CUDA-assembly-code. Moreover, the sass-instruction-set may change with the capabilities of GPUs, which increases the workload and reduces the usability. In contrast, full C++ implementations offers better applicability and readability.

*E. Integration to Dragon-Alpha & Cu32*

Dragon-Alpha [15][21], abbreviated as Alpha, is a Java-based Tensor Computing Framework, and can be used to express and execute deep learning algorithms. Cu32, a math library within Alpha, is designed for float32-operations on CUDA. Both Alpha and cu32 are fully developed by us. To run Alpha's applications, only an appropriate version of JDK and CUDA is required.

In this work, the kernel-functions of C-K-S have been merged to form a higher-level encapsulation in cu32, and have been integrated to Dragon-Alpha by using Java-Native-Interface. We have upgraded Alpha from version 1.0 to 1.1, resulting in about 10% increase in overall speed.

Dragon-Alpha adopts multiple methods to implement convolutional operators, including C-K-S, Winograd, implicit-GEMM, and direct-method. Each method has its own advantages, and is suitable for certain situations.

For deconvolution with a non-unit stride, KS-deconv is more efficient than the common approach. However, when stride is 1, the common approach is simpler and faster, due to the density of feature-maps.

The effect of filter-trimming is positively correlated with the size of zero-padding. Generally, filter-trimming is used, when the proportion of zero-padding exceeds a certain threshold (usually 6%). For instance, if the input-feature-maps are $(32\times32)$ and the padding is 1.

The input-feature-maps of many CNNs are much greater than $(32\times32)$, such as $(128\times128)$ and $(224\times224)$, where small zero-padding is not an issue to lower the performance. However, the size of feature-maps are easily reduced to $(32\times32)$ or smaller through down-sampling. Besides, convolutional operators applied to not-too-big feature-maps often have large channels, hence high complexity, where filter-trimming is commonly employed as a technique to address zero-padding.

For KS-deconv-V2, the input-feature-maps are smaller than the output-feature-maps. During the execution, the filters are split to multi smaller kernels, each of which is applied to a $\frac{1}{stride^2}$ subset of input-feature-maps. The subsets are zero-padded, and much smaller than the output-feature-maps, allowing more effective use of filter-trimming.

Filter-trimming reduces overhead in some aspects, but it may alter the sequence of memory-access, potentially decreasing the hit-ratio of the L2-cache. Additionally, other methods could be more effective, when there is large-feature-map, a limited number of zero-padding, or a big overlap between patches.

## V. EXPERIMENTS

This section examines the speed and convergence of C-K-S through two experiments. The first compares the speed of C-K-S on various sizes of feature-maps, against that of cuDNN [9]. The second experiment utilizes C-K-S to train VGG [3] and ResNet [5] on Cifar10 [24] and ILSVRC2012 [25] datasets, with comparison to PyTorch.

In both experiments, the CUDA version is 11.5, the cuDNN version is 8302, the PyTorch version is 1.12. All calculations were performed using float32 datatype. The major part of calculations was executed on GPU. To ensure optimal utilization of GPU and minimize interference from other programs, each program that executed an operator or a CNN was configured accordingly.

### A. Experiment1: Speed in different scenarios

*1) Methods and Conditions:* In this experiment, cuDNN serves as the benchmark for comparison, due to its highly optimized convolutional operators. These operators were utilized through PyTorch, since cuDNN has been integrated into PyTorch as an underlying library.

We consider convolution, deconvolution and dilated-convolution as a collective conv-layer. The evaluation of convolution operators focused on the forward propagation denoted by (1), while deconvolution and dilated-convolution operators were evaluated in terms of backward propagation represented by (2) and (3). Two sets of test-cases were provided, for each of convolution, deconvolution, and dilated-convolution. Each set has unique size of filters, and includes 8 test cases. The size of feature-maps decreases from the 1$^{st}$ to the 8$^{th}$ test case, while the channels and batch-size increase to ensure a sufficient large data-size for GPU parallelism.

Each operation corresponding to a test case was executed 1000 times consecutively on an RTX3060ti GPU. The operations performed using C-K-S and PyTorch were logically equivalent, resulting in the same time-complexity. The completion of each execution was ensured by using *cudaDeviceSynchronize*(). The total execution time was averaged, and the speed was calculated based on the time-complexity and average time. The speed is measured in GFlop/s ($10^9$ float32 operations per second). The time-complexity formulas of convolutional operators can be found in Table III.

To maximize PyTorch's speed, sufficient memory was pre-allocated before executing a loop of a convolutional operator. PyTorch didn't perform any transpositions or rotations to change the arrangement of tensors, but solely executed convolutional operators.

TABLE III. FORMULAS FOR TIME-COMPLEXITY

| Formula | Explanation |
|---|---|
| $T_{Conv} = 2 * (O_C * N * O_H * O_W * F_H * F_W * I_C)$ | The time-complexity of 2D-Convolution. |
| $T_{Deconv} = 2 * (I_C * N * I_H * I_W * F_H * F_W * O_C)$ | The time-complexity of 2D-Deconvolution. |
| $T_{Dilated} = 2 * (O_C * F_H * F_W * I_C * O_H^p * O_W^p)$ | The time-complexity of 2D-Dilated-Convolution. |
| $O_H^p = O_H + (O_H - 1) * (sh - 1)$ | The height of $\nabla Y$ with inserted $(sh - 1)$ 0s. |
| $O_W^p = O_W + (O_W - 1) * (sw - 1)$ | The width of $\nabla Y$ with inserted $(sw - 1)$ 0s. |

*2) Results and Discussions:* Six line-charts are used to illustrate the speed-variation under different tensor shapes. Figures 9-10 display the results of convolution, where the symbol '*' denotes the exclusion of filter-transposition time. Figures 11-12 present the results of deconvolution, while Figures 13-14 list the results of dilated-convolution.

Based on the deconvolution results, KS-deconv-V2 outperforms PyTorch in all 16 test-cases. According to the results of convolution and dilated-convolution, PyTorch is faster in the middle test-cases, while the opposite is observed in the other test-cases. In CNNs, conv-layers with big channels and medium feature-maps, usually have the highest time-complexity. The middle test-cases are representatives of these conv-layers. Therefore, cuDNN has been fully optimized at assembly level for these cases, to be more efficient than our implementations.

In this experiment, the proportion of padded 0s is bigger in smaller feature-maps. These 0s disrupt the SIMD of cuDNN, but allow filter-trimming to play a more significant role. As a result, the speed of ConvV2 and Sk-dilated-V2 ultimately surpasses that of PyTorch. Besides, the filter-trimming plays greater role in the execution of KS-deconv, leading to a higher acceleration.

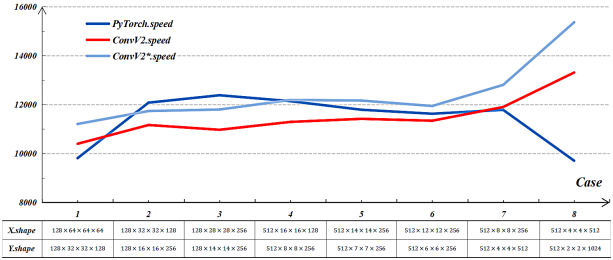

**Fig. 9.** Convolution: $\langle F_H \ F_W \rangle = \vec{3}, padding = 1$.

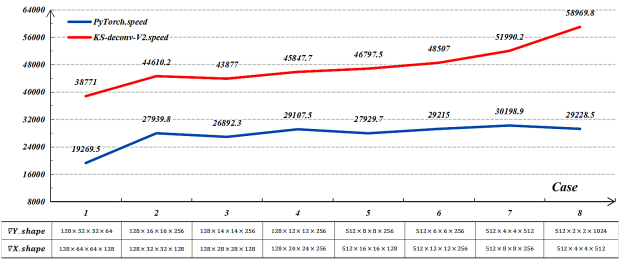

**Fig. 11.** Deconvolution: $\langle F_H \ F_W \rangle = \vec{3}, padding = 1$.

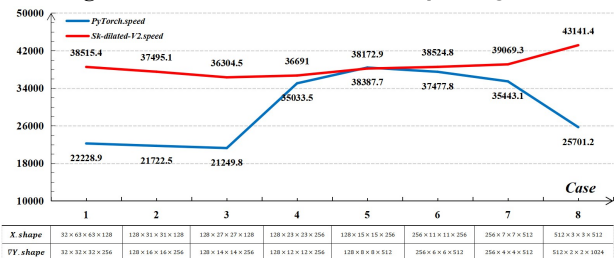

**Fig. 13.** Dilated-Convolution: $\langle F_H \ F_W \rangle = \vec{3}, padding = 1$.

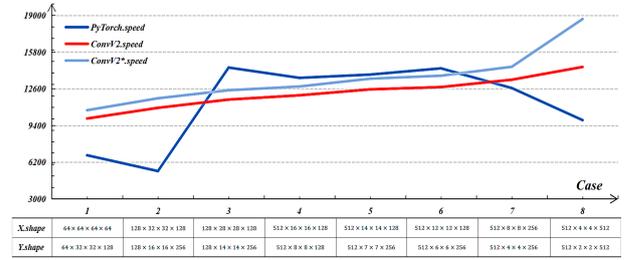

**Fig. 10.** Convolution: $\langle F_H \ F_W \rangle = \vec{5}, padding = 2$.

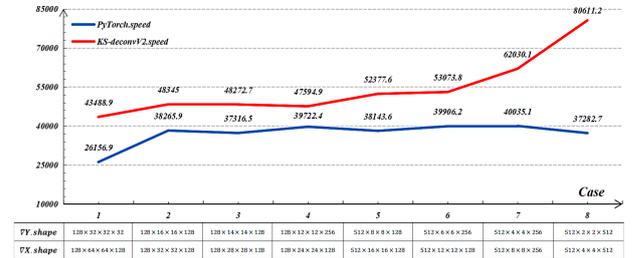

**Fig. 12.** Deconvolution: $\langle F_H \ F_W \rangle = \vec{5}, padding = 2$.

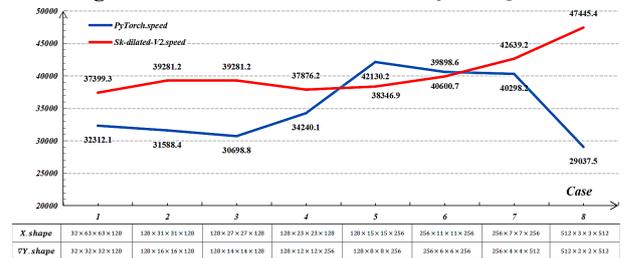

**Fig. 14.** Dilated-Convolution: $\langle F_H \ F_W \rangle = \vec{5}, padding = 2$.

In the context of deconvolution and dilated-convolution, tensor $\nabla Y$ expands nearly 4 times due to zero-insertion. KS-deconv and Sk-dilated skip these 0s, resulting in two enhancements. First, the actual time-complexity is only about 25% of the theoretical value. Second, there is no sparse tensor involved in calculation, meaning that some design patterns of ConvV2 can be applied to KS-deconv and Sk-dilated. This enables KS-deconv and Sk-dilated utilize the GPU to a similar extent as ConvV2. These two enhancements are reasons why the speed of KS-deconv and Sk-dilated is close to 4 times that of ConvV2's. However, it's difficult to surpass this '4 times', as the cost of KS-deconv and Sk-dilated is higher than ConvV2. KS-deconv requires additional resources for filter-reconstruction, and more registers for intermediate variables, which reduces the parallelism of computationally-intensive kernel-functions. Sk-dilated relies on map-reduce to ensure parallelism, that incurs certain expenses. Compared to ConvV2, its memory-access is less continuous with larger spans, leading to a lower hit ratio of L2-cache.

For ConvV2 or KS-deconv, about 5% of the overall computing time is dedicated to filter-transposition or filter-construction. These two filter-operations can be hidden in the parallel execution of multi operators. When conducting one-way propagation, these two filter-operations can be performed prior to loops of propagation, so that the total calculation time can be further reduced.

In conclusion, this experiment and analysis provide evidence supporting the efficacy of C-K-S.

### B. Experiment2: training of CNNs

To analyze the speed and convergence of C-K-S, we integrated C-K-S to Dragon-Alpha, and used Alpha to train several typical CNNs, on Cifar10 and ILSVRC2012 datasets, with comparison to PyTorch.

*1) Methods and Conditions:* The CNNs of Alpha and PyTorch, were constructed with identical logic, and were subjected to the same initialization, training and testing conditions. The CNNs were trained by Adam [26], with specific hyper-parameters $learning\_rate = 0.001$, $\beta_1 = 0.9$, $\beta_2 = 0.999$, and $\varepsilon = 10^{-8}$. In order to expedite convergence, BatchNorm [27] was applied in VGG. Full-connect layers and conv-layers were initialized using kaiming-uniform [28]. The weight and bias of BatchNorm layers are respectively initialized to 1 and 0.

For Cifar10, the CNNs are trained on an RTX3060ti GPU, the label is 10-class, the shape of input is $(32 \times 32 \times 3)$, and the batch-size is 512. For ILSVRC2012, the CNNs are trained on an RTX4090 GPU, the label is 1000-class, the shape of input is $(128 \times 128 \times 3)$, and the batch-size is 256.

For both datasets, labels are transformed to a one-hot format, and the pixel values of inputs are linearly scaled between $-1$ and $+1$. Specific full-connect layers are modified to accommodate the shape of tensors.

The loss-function value was recorded every 10 steps, to generate a loss-sequence. For Cifar10, each CNN was trained for 5 times to obtain 5 loss-sequences. These 5 sequences were averaged to draw the loss-function curve. For ILSVRC2012, each CNN was trained once, because of the lengthy training process. To draw the loss-function curve, We averaged the 10 adjacent elements in the loss-sequence without any overlap. Additionally, we used 'nvidia-smi' to calculate GPU-memory-usage.

*2) Results and Discussions:* The performance of Alpha and PyTorch on Cifar10 is displayed in Table IV, while the loss-function curves can be observed in Figures 15-18. For ILSVRC2012, the performance is presented in Table V, and the loss-functions curves are depicted in Figures 19-22.

On both datasets, the CNNs trained using Alpha and

TABLE IV. Performance of Alpha and PyTorch on Cifar10

| Network | Speed | Acceleration | Accuracy on train \ test set | GPU-Memory-usage | Weight-File-Size |
|---|---|---|---|---|---|
| ResNet18 (trained 25 epoch) | Alpha: 5.822 s / epoch<br>Torch: 7.312 s / epoch | 1.256x | Alpha: 99.09%  78.11%<br>Torch: 98.90%  77.90% | Alpha: 1067 MB<br>Torch: 2486 MB | Alpha: 66.7 MB<br>Torch: 48.2 MB |
| ResNet34 (trained 30 epoch) | Alpha: 11.528 s / epoch<br>Torch: 14.149 s / epoch | 1.227x | Alpha: 99.01%  79.45%<br>Torch: 98.87%  79.16% | Alpha: 1685 MB<br>Torch: 3116 MB | Alpha: 120 MB<br>Torch: 87.3 MB |
| VGG16 (trained 35 epoch) | Alpha: 9.709 s / epoch<br>Torch: 11.041 s / epoch | 1.137x | Alpha: 97.91%  82.75%<br>Torch: 97.59%  82.62% | Alpha: 1634 MB<br>Torch: 3658 MB | Alpha: 78.7 MB<br>Torch: 56.7 MB |
| VGG19 (trained 40 epoch) | Alpha: 11.849 s / epoch<br>Torch: 13.142 s / epoch | 1.109x | Alpha: 96.06%  81.13%<br>Torch: 96.03%  80.98% | Alpha: 1731 MB<br>Torch: 3740 MB | Alpha: 106 MB<br>Torch: 77 MB |

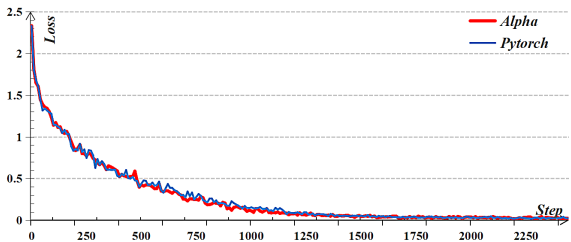

**Fig. 15.** Cifar10: ResNet18 (trained 25 epoch).

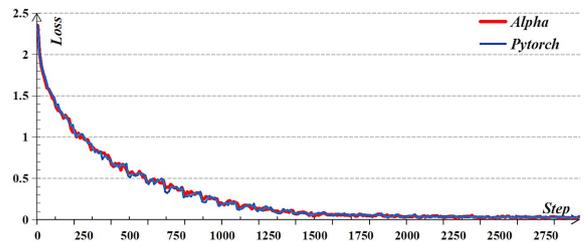

**Fig. 16.** Cifar10: ResNet34 (trained 30 epoch).

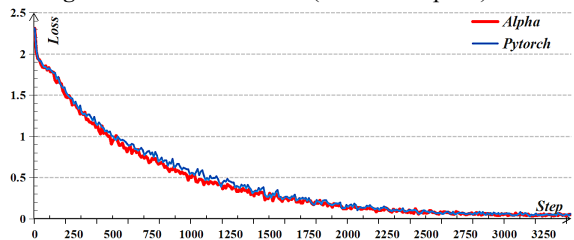

**Fig. 17.** Cifar10: VGG16 (trained 35 epoch).

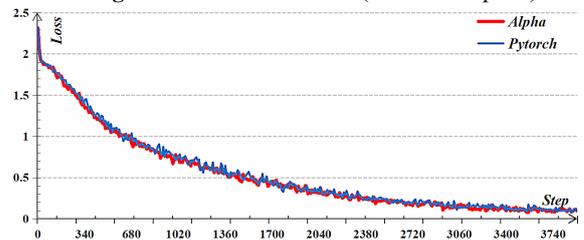

**Fig. 18.** Cifar10: VGG19 (trained 40 epoch).

TABLE V. Performance of Alpha and PyTorch on ILSVRC2012

| Network | Speed | Acceleration | accuracy on train set | GPU-Memory-usage | Weight-File-Size |
|---|---|---|---|---|---|
| ResNet18 (trained 50 epoch) | Alpha: 664.957 s / epoch<br>Torch: 1202.894 s / epoch | 1.809x | Alpha: 97.99%<br>Torch: 98.30% | Alpha: 5718 MB<br>Torch: 11848 MB | Alpha: 66.8 MB<br>Torch: 50.9 MB |
| ResNet34 (trained 50 epoch) | Alpha: 1264.301 s / epoch<br>Torch: 2219.791 s / epoch | 1.756x | Alpha: 98.27%<br>Torch: 98.38% | Alpha: 9090 MB<br>Torch: 15384 MB | Alpha: 124 MB<br>Torch: 89.8 MB |
| VGG16 (trained 30 epoch) | Alpha: 1087.803 s / epoch<br>Torch: 1115.235 s / epoch | 1.025x | Alpha: 97.94%<br>Torch: 97.65% | Alpha: 10870 MB<br>Torch: 14240 MB | Alpha: 294 MB<br>Torch: 223 MB |
| VGG19 (trained 40 epoch) | Alpha: 1217.609 s / epoch<br>Torch: 1246.722 s / epoch | 1.024x | Alpha: 97.85%<br>Torch: 97.42% | Alpha: 11138 MB<br>Torch: 14438 MB | Alpha: 319 MB<br>Torch: 244 MB |

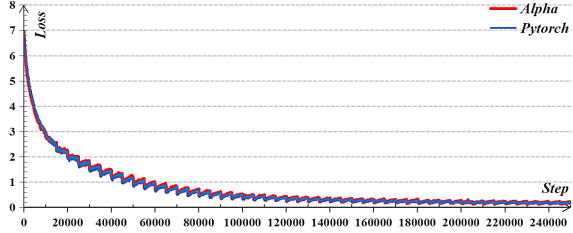

**Fig. 19.** ILSVRC2012: ResNet18, 50 epoch.

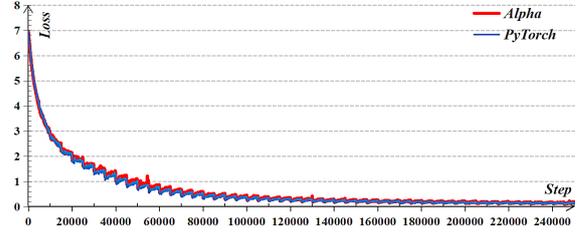

**Fig. 20.** ILSVRC2012: ResNet34, 50 epoch.

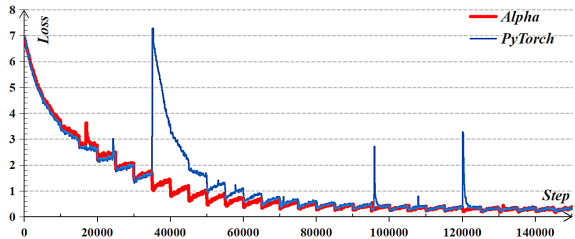

**Fig. 21.** ILSVRC2012: VGG16, 30 epoch.

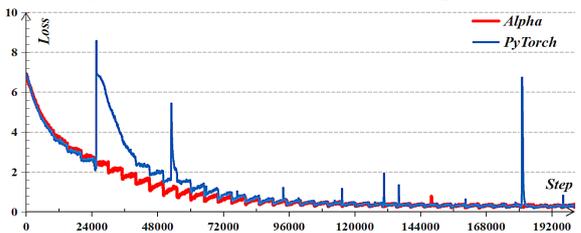

**Fig. 22.** ILSVRC2012: VGG19, 40 epoch.

PyTorch demonstrate comparable trends in loss-function curves and accuracy. However, Alpha outperforms PyTorch in terms of speed and GPU-memory-usage. Notably, when training VGG using PyTorch on ILSVRC2012, there are fluctuations in the loss-functions curves, primarily attributed to an excessively high learning-rate of 0.001. As C-K-S plays a crucial role in training CNNs, these experimental results are evidence supporting its correctness and efficiency.

The low memory-usage of Alpha further supports the cost-effectiveness of C-K-S in relation to its auxiliary memory needs; although the low-memory-usage can also be attributed to Alpha's memory-management-mechanisms, inplace-operators and fused-operators.

In contrast to ResNet, VGG utilizes max-pooling for down-sampling rather than non-unit convolution, which restricts the potential contributions of KS-deconv and Sk-dilated. That's one of the reasons why Alpha achieves lower acceleration on VGG than ResNet.

In comparison to ILSVRC2012, Cifar10 has smaller feature-maps, allowing for more reduction of overhead in VGG training through filter-trimming. As a result, Alpha achieves higher acceleration of VGG on Cifar10.

## VI. CONCLUSION

This paper discusses the background, motivations, algorithms and implementations of the C-K-S algorithm. During this work, we developed high-performance GPU implementations of C-K-S, and integrated them to Dragon-Alpha [15] to train CNNs. The correctness and efficiency of C-K-S has been demonstrated through two experiments, with comparison to cuDNN [9] and PyTorch [13].

The C-K-S algorithm skips 0-calculations through filter-trimming, filter-reconstruction and leaping-access. C-K-S not only reduces the computational complexity of conv-layers, but also simplifies the control of hardware to make acceleration easier. As a result, C-K-S has the potential to surpass the capabilities of existing devices.

The higher-dimensional versions of the C-K-S can be analogized to its 2D counterpart discussed in this paper. Further implementations of C-K-S can be based on CPU, FPGA, TPU or other hardware.

# APPENDIX

**Algorithm. 1** *ConvV2*

**Input:** input-features $X \subset \mathbf{R}^{N \times I_H \times I_W \times I_C}$,
 filters $W \subset \mathbf{R}^{O_C \times F_H \times F_W \times I_C}$,
 stride $\langle sh\ sw \rangle$, padding $\langle ph\ pw \rangle$
**Output:** output-features $Y \subset \mathbf{R}^{N \times O_H \times O_W \times O_C}$

**for** $Y_{n,oh,ow,oc}$ **in** $Y$
 $\langle ih_s\ iw_s \rangle = \langle oh, ow \rangle \odot \langle sh\ sw \rangle - \langle ph\ pw \rangle$
 $\begin{pmatrix} fh_s & fh_e \\ fw_s & fw_e \end{pmatrix} = \begin{pmatrix} max(-ih_s, 0) & min(I_H - ih_s,\ F_H) \\ max(-iw_s, 0) & min(I_W - iw_s,\ F_W) \end{pmatrix}$
 $Y_{n,oh,ow,oc} = \sum_{\substack{fh \in [fh_s, fh_e] \\ fw \in [fw_s, fw_e] \\ ic \in [0, I_C)}} X_{n,(ih_s+fh),(iw_s+fw),ic} * W_{oc,fh,fw,ic}$

---

**Algorithm. 2** *KS-deconv:Stage1*

**Input:** filters $W \subset \mathbf{R}^{O_C \times F_H \times F_W \times I_C}$,
 stride $\langle sh\ sw \rangle$
**Output:** the smaller kernels $C \subset \mathbf{R}^{sh \times sw \times O_C \times \lceil \frac{F_H}{sh} \rceil \times \lceil \frac{F_W}{sw} \rceil \times I_C}$
**Initialize:** $C = 0$

**for** $C_{y,x}$ **in** $C$
 $\langle oph_{y,x}\ opw_{y,x} \rangle = \langle \lceil \frac{F_H - y}{sh} \rceil\ \lceil \frac{F_W - x}{sw} \rceil \rangle - \vec{1}$
 **for** $C_{y,x,oc,ch,cw}$ **in** $C_{y,x}$
  $\langle fh\ fw \rangle = \langle y\ x \rangle + \langle oph_{y,x} - ch\ opw_{y,x} - cw \rangle \odot \langle sh\ sw \rangle$
  $C_{y,x,oc,ch,cw} = W_{oc,fh,fw}$ **if** $\langle fh\ fw \rangle$ **in-range-of** $W$

---

**Algorithm. 2** *KS-deconv:Stage2&3*

**Input:** gradient of output-features $\nabla Y \subset R^{N \times O_H \times O_W \times O_C}$,
 the smaller kernels $C = 0 \subset \mathbf{R}^{sh \times sw \times O_C \times \lceil \frac{F_H}{sh} \rceil \times \lceil \frac{F_W}{sw} \rceil \times I_C}$
 stride $\langle sh\ sw \rangle$, padding $\langle ph\ pw \rangle$
**Output:** gradient of input-features $\nabla X \subset \mathbf{R}^{N \times I_H \times I_W \times I_C}$

**for** $C_{y,x}$ **in** $C$
 $\langle ih_s\ iw_s \rangle = \langle y\ x \rangle - \langle ph\ pw \rangle$
 $\langle ih_s\ iw_s \rangle\ += \langle 1_{ih_s<0}\ 1_{iw_s<0} \rangle \odot \langle \lceil \frac{-ih_s}{sh} \rceil\ \lceil \frac{-iw_s}{sw} \rceil \rangle \odot \langle sh\ sw \rangle$
 **for** $\langle n\ u\ v\ ic \rangle = \vec{0}$ **to** $\langle N\ \lceil \frac{I_H}{sh} \rceil\ \lceil \frac{I_W}{sw} \rceil\ I_C \rangle - \vec{1}$
  $\langle ih\ iw \rangle = \langle u\ v \rangle \odot \langle sh\ sw \rangle + \langle ih_s\ iw_s \rangle$
  $\langle oh_s\ ow_s \rangle = \langle \lceil \frac{ih+ph-y}{sh} \rceil\ \lceil \frac{iw+pw-x}{sw} \rceil \rangle - \langle oph_{y,x}\ opw_{y,x} \rangle$
  **continue if** $\langle ih\ iw \rangle$ **not in-range-of** $\nabla X$
  $\nabla X_{n,iw,ih,ic} = 0$
  **for** $C_{y,x,oc,ch,cw,ic}$ **in** $C_{y,x}$:
   $\langle oh\ ow \rangle = \langle oh_s\ ow_s \rangle + \langle ch\ cw \rangle$
   **continue if** $\langle oh\ ow \rangle$ **not in-range-of** $\nabla Y$
   $\nabla X_{n,ih,iw,ic}\ += \nabla Y_{n,oh,ow,oc} * C_{y,x,oc,ch,cw,ic}$

---

**Algorithm. 2B** *KS-deconv-V2:Stage2&3*

**Input:** gradient of output-features $\nabla Y \subset R^{N \times O_H \times O_W \times O_C}$,
 the smaller kernels $C = 0 \subset \mathbf{R}^{sh \times sw \times O_C \times \lceil \frac{F_H}{sh} \rceil \times \lceil \frac{F_W}{sw} \rceil \times I_C}$
 stride $\langle sh\ sw \rangle$, padding $\langle ph\ pw \rangle$
**Output:** gradient of input-features $\nabla X \subset \mathbf{R}^{N \times I_H \times I_W \times I_C}$

**for** $C_{y,x}$ **in** $C$
 $\langle C_H^{y,x}\ C_W^{y,x} \rangle = \langle \lceil \frac{F_H - y}{sh} \rceil\ \lceil \frac{F_W - x}{sw} \rceil \rangle$
 $\langle ih_s\ iw_s \rangle = \langle y\ x \rangle - \langle ph\ pw \rangle$
 $\langle ih_s\ iw_s \rangle\ += \langle 1_{ih_s<0}\ 1_{iw_s<0} \rangle \odot \langle \lceil \frac{-ih_s}{sh} \rceil\ \lceil \frac{-iw_s}{sw} \rceil \rangle \odot \langle sh\ sw \rangle$
 **for** $\langle n\ u\ v\ ic \rangle = \vec{0}$ **to** $\langle N\ \lceil \frac{I_H}{sh} \rceil\ \lceil \frac{I_W}{sw} \rceil\ I_C \rangle - \vec{1}$
  $\langle ih\ iw \rangle = \langle u\ v \rangle \odot \langle sh\ sw \rangle + \langle ih_s\ iw_s \rangle$
  $\langle oh_s\ ow_s \rangle = \langle \lceil \frac{ih+ph-y}{sh} \rceil\ \lceil \frac{iw+pw-x}{sw} \rceil \rangle - \langle oph_{y,x}\ opw_{y,x} \rangle$
  $\begin{pmatrix} ch_s & ch_e \\ cw_s & cw_e \end{pmatrix} = \begin{pmatrix} max(-oh_s, 0) & min(O_H - oh_s,\ oph_{y,x}) \\ max(-ow_s, 0) & min(O_W - ow_s,\ opw_{y,x}) \end{pmatrix}$
  **continue if** $\langle ih\ iw \rangle$ **not in-range-of** $\nabla X$
  $\nabla X_{n,ih,iw,ic} = \sum_{\substack{ch \in [ch_s, ch_e] \\ cw \in [cw_s, cw_e] \\ ic \in [0, I_C)}} \nabla Y_{n,(oh_s+ch),(ow_s+cw),oc} * C_{y,x,oc,ch,cw,ic}$

---

**Algorithm. 3** *Sk-dilated*

**Input:** gradient of output-features $\nabla Y \subset \mathbf{R}^{N \times OH \times OW \times OC}$,
 input-features $X \subset \mathbf{R}^{N \times IH \times IW \times IC}$,
 stride $\langle sh\ sw \rangle$, padding $\langle ph\ pw \rangle$
**Output:** gradient of filters $\nabla W \subset \mathbf{R}^{OC \times FH \times FW \times IC}$

**for** $\nabla W_{oc,fh,fw,ic}$ **in** $\nabla W$
 $\langle ih_s\ iw_s\ \nabla W_{oc,fh,fw,ic} \rangle = \langle fh - ph\ fw - pw\ 0 \rangle$
 **for** $\nabla Y_{n,oh,ow,oc}$ **in** $\nabla Y$
  $\langle ih\ iw \rangle = \langle oh\ ow \rangle \odot \langle sh\ sw \rangle + \langle ih_s\ iw_s \rangle$
  **continue if** $\langle ih\ iw \rangle$ **not in-range-of** $X$
  $\nabla W_{oc,fh,fw,ic}\ += X_{n,ih,iw,ic} * \nabla Y_{n,oh,ow,oc}$

---

**Algorithm. 3B** *Sk-dilated-V2*

**Input:** gradient of output-features $\nabla Y \subset \mathbf{R}^{N \times OH \times OW \times OC}$,
 input-features $X \subset \mathbf{R}^{N \times IH \times IW \times IC}$,
 stride $\langle sh\ sw \rangle$, padding $\langle ph\ pw \rangle$
**Output:** gradient of filters $\nabla W \subset \mathbf{R}^{OC \times FH \times FW \times IC}$

**for** $\nabla W_{oc,fh,fw,ic}$ **in** $\nabla W$
 $\langle ih_s\ iw_s \rangle = \langle fh - ph\ fw - pw \rangle$
 $\begin{pmatrix} oh_s & oh_e \\ ow_s & ow_e \end{pmatrix} = \begin{pmatrix} max(\lceil \frac{-ih_s}{sh} \rceil, 0) & min(O_H,\ \lceil \frac{I_H - ih_s}{sh} \rceil) \\ max(\lceil \frac{-iw_s}{sw} \rceil, 0) & min(O_W,\ \lceil \frac{I_W - iw_s}{sw} \rceil) \end{pmatrix}$
 $\nabla W_{oc,fh,fw,ic} = \sum_{\substack{oh \in [oh_s, oh_e] \\ ow \in [ow_s, ow_e] \\ n \in [0, N)}} X_{n,(ih_s+oh*sh),(iw_s+ow*sw),ic} * \nabla Y_{n,oh,ow,oc}$